\documentclass[a4paper,conference]{IEEEtran}
\usepackage[hyphens]{url}  
\usepackage{graphicx} 
\usepackage{caption} 
\DeclareCaptionStyle{ruled}{labelfont=normalfont,labelsep=colon,strut=off} 
\usepackage{soul}
\usepackage{subfigure}
\usepackage{mathtools}
\usepackage{amssymb}
\usepackage{amsmath}
\usepackage{amsthm}
\usepackage{algorithm}
\usepackage{multirow}
\usepackage{lineno}
\usepackage{xcolor}
\usepackage{algpseudocode}
\usepackage{cite}
\usepackage{pdfpages}
\newcommand{\algrule}[1][.2pt]{\par\vskip.5\baselineskip\hrule height #1\par\vskip.5\baselineskip}
\newtheorem{definition}{Definition}
\newtheorem*{example*}{Example}

\begin{document}
\title{\textsc{AutoCoMet}: Smart Neural Architecture Search via Co-Regulated Shaping Reinforcement}


\author{\IEEEauthorblockN{Mayukh Das\IEEEauthorrefmark{1},
Brijraj Singh\IEEEauthorrefmark{2},
Harsh Kanti Chheda\IEEEauthorrefmark{3},
Pawan Sharma\IEEEauthorrefmark{4} and
Pradeep NS\IEEEauthorrefmark{5}}
\IEEEauthorblockA{\IEEEauthorrefmark{1}Microsoft Research; \IEEEauthorrefmark{2}Sony Research; \IEEEauthorrefmark{3}JP Morgan \& Chase; \IEEEauthorrefmark{4}Microsoft; \IEEEauthorrefmark{5}Samsung R\&D Institute, Bangalore}
\IEEEauthorblockA{Email: \IEEEauthorrefmark{1}mayukhdas@microsoft.com; \IEEEauthorrefmark{2}brijraj.singh@sony.com; \IEEEauthorrefmark{5}pradeep.ns@samsung.com}}

\maketitle

\begin{abstract}
Designing suitable deep model architectures, for AI-driven on-device apps and features, at par with rapidly evolving mobile hardware and increasingly complex target scenarios is a difficult task. 
Though Neural Architecture Search (NAS/AutoML) has made this easier by shifting paradigm from extensive manual effort  to automated architecture learning from data, 
yet it has major limitations, leading to critical bottlenecks in the context of mobile devices, including model-hardware fidelity, prohibitive search times and deviation from primary target objective(s). 
Thus, we propose \textsc{AutoCoMet} that can learn the most suitable DNN architecture optimized for varied types of device hardware and task contexts, $\approx 3\times$ faster. Our novel co-regulated shaping reinforcement controller together with the high fidelity hardware meta-behavior predictor produces a smart, fast NAS framework that adapts to context via a generalized formalism for any kind of multi-criteria optimization.  
\end{abstract}

\section{Introduction}

The ability to deploy AI models, especially Deep Neural Networks (DNNs) on resource-constrained devices (smart phones/tabs/TVs and edge sensors) has transformed the way these devices are used. While increasingly sophisticated deep models empower intelligent features and apps, revolutionary hardware and/or neural acceleration platforms \cite{snpe,snap,coreml} allow us to enable such features on even smaller devices. 

However, extensive manual effort by AI experts/professionals is needed for model architecture design and hyperparameter control in order to be effective and efficient for a given hardware and task scenario \cite{bergstra2012random}. This is even more critical in context of embedded AI. \textit{For instance}, large  transformer models (GPT-3 etc.) are indispensable for text prediction, but needs sparse parameters, shorter embedding lengths or limited attention heads to work on low-resource devices. \textit{Similarly MobileNet-V2}, a breakthrough in device-efficient vision model via `separable convolutions', may be hard to deploy on lower-end devices unless several architectural parameters are adjusted. 
Manual design of DNN architectures of every new case is not sustainable in an evolving landscape of device hardware and AI-driven scenarios.
Neural Architecture Search (NAS) or AutoML
\cite{liu2018progressive,elsken2019neural,xie2018snas} 
allows for automatic learning (fine-tuning) suitable architectures from data (and/or additional information) and has shown a lot of promise \cite{tan2019mnasnet,cai2018proxylessnas}. 

But, NAS is largely \textit{not 'smart'} enough to adapt to evolving hardware configurations and increasingly complex task and optimality requirements (jointly termed as context here). There are major challenges. 
(1) Powerful NAS-designed device-efficient models (ex: EfficientNet\cite{tan2019efficientnet}), may be highly optimized for certain h/w (ex: Google TPUs) but may struggle on other chipsets or acceleration platforms. DNNs learned by NAS for high-end smart devices may either be fully/partially incompatible or inefficient in older/newer or low-tier devices. (2) Improper design of objectives can lead NAS to compress models so aggressively that they suffer in predictive performance or other factors. (3) Again, models designed for one task (such as object detection) may not generalize well to other related tasks (such as caption detection). (4) Most importantly, even robust NAS frameworks that arguably try to address these, have a major bottleneck - prohibitive search time that makes industrial adaptation of NAS extremely challenging. 

To address these challenges we propose our smart context-adaptive generalized NAS framework, \textbf{\underline{Auto}}ML via \textbf{\underline{Co}}-Regulated \textbf{\underline{Met}}a-behavior  (\textsc{AutoCoMet}), that can automatically generate / customize and learn most suitable DNN architecture as per the context. `Context' jointly signifies varied levels of hardware configurations and task/quality requirements. Not only does our framework provide a generalized formalism of the multi-criteria objective but also proposes a novel way to solve the same via Reinforcement Learning with \underline{co-regulated reward shaping} resulting in rapid search. 

We make the following contributions: \textbf{(1)} We propose a context-adaptive NAS/AutoML framework, \textbf{(2)} To that end we propose a high fidelity hybrid ensemble meta-behavior predictor that can effectively handle mixed feature space producing accurate hardware behaviour prediction, \textbf{(3)} We propose a novel co-regulated reward shaping Reinforcement Learning (RL) NAS controller that that leads generalization in a principled manner as well as extremely rapid search and \textbf{(4)} Finally, we demonstrate empirically in several important scenarios how our framework is effective in  generating suitable neural architectures for varied contexts.

\section{Background \& Related Work}
\subsection{Neural Architecture Search / AutoML}
Although the spirit remains same, there are several classes of NAS methodologies depending on multiple dimensions. 
\paragraph{Search strategy}Approaches may differ in their search strategies - \textbf{(1)} Reinforcement Learning (zero-order optimization) based methods \cite{zoph2016neural, tan2019mnasnet, jaafra2019reinforcement}; \textbf{(2)} First-order optimization or gradient-based search including UNAS, DARTS  etc., \cite{unas,liu2018darts} or Bayesian approaches like BayesNAS~\cite{zhou2019bayesnas}; \textbf{(3)} Evolutionary algorithms, known to be powerful in  combinatorial multi-objective search in the discrete space (eg.  LEMONADE \cite{elsken2018efficient} etc). \textit{We choose to adopt a zero-order Reinforcement Learning formalism since we mostly operate in a discrete design space with potentially piece-wise metrics, and hence differentiability is not trivial.}
\paragraph{Design space and search latency}Several approaches try to address the concerns of intractable {search space} and search latency. Ex: ENAS \cite{pmlr-v80-pham18a} leverages parameter sharing among all tunable  intermediate neural architectures. LEMONADE \cite{elsken2018efficient} proposes Lamarckian based inheritance to handle the resource expensive nature of the NAS controller. Liu et al., \cite{liu2018progressive} progressively searches complex architectures to keep the design space controlled. Recently, HourNAS \cite{hournas} mitigates search complexity by identification and added attention on vital blocks in a deep model. 
\textit{Our work proposes a novel way to navigate/explore the combinatorial search space instead and consequently helps control search latency.}

\paragraph{On-device models}Inference efficiency of learned architectures is critical for on-device execution of DNNs, as inference entails complex tensor computations and affects user experience. MNASNet \cite{tan2019mnasnet} was one of the first ones to optimize architectures for mobile devices using a factored hierarchical search space 
and used a simple multi-criteria RL controller. Many other approaches also focus on hardware-adaptive search to find latency-optimal architectures for resource-limited devices \cite{fbnet,cai2018proxylessnas,xie2018snas,unas,chamnet}. Several state-of-the-art vision models for embedded AI were learned with NAS \cite{howard2019searching,tan2019efficientnet}.  Our focus, as well, is to design a NAS framework for on-device deep models. However, we aim to make it smart, context-adaptive, generalized and fast. 

ProxylessNAS \cite{cai2018proxylessnas} is probably the closest in spirit to our work. Like MNASNet, they adopt a weighted product formulation of the multi-objective problem and try to use direct hardware level execution statistic (ex: latency) instead of complexity proxies. They use 2 parallel search strategies -  Gradient-based and REINFORCE \cite{williams1992simple}. But, ProxylessNAS (and MNASNet) has some major limitations including - (1) limited latency estimator for mobiles; (2) naive (weighted-combination) strategy for multi-criteria RL and (3) stateless RL formulation which leads to predecessor-successor dependencies not being captured faithfully. \textsc{AutoCoMet} addresses such limitations with high-fidelity meta-behavior predictor, novel generalized multi-criteria optimizer using co-regulated reward-shaping RL.  

\subsection{Augmented Reinforcement Learning}
As hinted earlier, \textsc{AutoCoMet} adopts an RL-based search, since representing DNN architecture design space as a sequence of state transitions restricts the complexity and a zero-order optimization scheme is useful in such a hybrid space. However a critical problem with RL, especially in our problem context, is `credit assignment'/feedback sparsity \cite{sutton1985temporal,fu2008solving,agogino2004unifying}.Architecture design space is potentially infinitely large and most architecture altering actions may not fetch any significant reward/feedback. Also this is a multi-criteria problem  \cite{gabor1998multi,van2014multi,liu2014multiobjective} and most existing approaches resort to scalarization/combination of reward vector which does not work in our scenario. 
Augmented RL, such as Reward-shaping \cite{grzes2017reward,grzes2008plan,ng1999policy}, is one of the ways to address feedback sparsity where extra inputs (knowledge) augments the original sprase reward space while ensuring policy invariance. Our formulation uses reward shaping to address feback sparsity in a complex multi-criteria space.  

\section{The \textsc{AutoCoMet} framework}
\subsection{Problem Setting}

Most NAS approaches have limited generalizability across different classes of hardware/task. This problem is more critical in mobile devices where variations in hardware configuration affects user experience of AI-driven features. 
This is due to certain limitations in existing AutoML approaches --  
    \textbf{(1)} {Adapting the search space} in an evolving mobile hardware and task landscape is non-trivial; 
    \textbf{(2)} {Naive combination} of the dimensions of a multi-criteria objective (scalarization) and/or {expensive combinatorial search} leading to prohibitive search latency; 
    \textbf{(3)} {Formalism is not general}. Some NAS approaches optimize latency too aggressively making accuracy suffer or vice versa. Manually designing this trade-off is difficult. Also the dimensions in a multi-criteria objective cannot be easily extended.
Hence our strategy enables suitable generalization across hardware/task and quality requirements (`Context'). 

\noindent \fbox{
\parbox{0.97\columnwidth}{
\textbf{GIVEN:} Context information (hardware configuration details + task/quality requirements).\\
\textbf{TO DO:} Develop a smart fast context-adaptive generalized NAS framework to learn the most suitable DNN arch.}}

\subsection{Framework Architecture}
\textit{``Smart'', in cognitive intelligence, is usually associated with fast learning, adapting to contexts and situations, and the ability to reason along multiple dimensions without losing sight of what is important. }
Our proposed framework, \textbf{\underline{Auto}}ML via \textbf{\underline{Co}}-regulated \textbf{\underline{Met}}a-behavior  (\textsc{AutoCoMet}) can hence be considered ``smart''. It has 2 major components, 
    \textbf{[I]} High-Fidelity Context meta-behavior prediction model that is trained from DNN execution stats from a wide range of devices. This results in a piece-wise differentiable behavior function that feeds one or more of the reward components; 
    \textbf{[II]} Novel multi-criteria Reinforcement Learning (RL) NAS controller that leverages a reward-shaping based `co-regulated' update strategy which leads to faster convergence. 
\begin{figure*}[h]
    \begin{minipage}{0.6\textwidth}
    \centering
    \includegraphics[height=4cm]{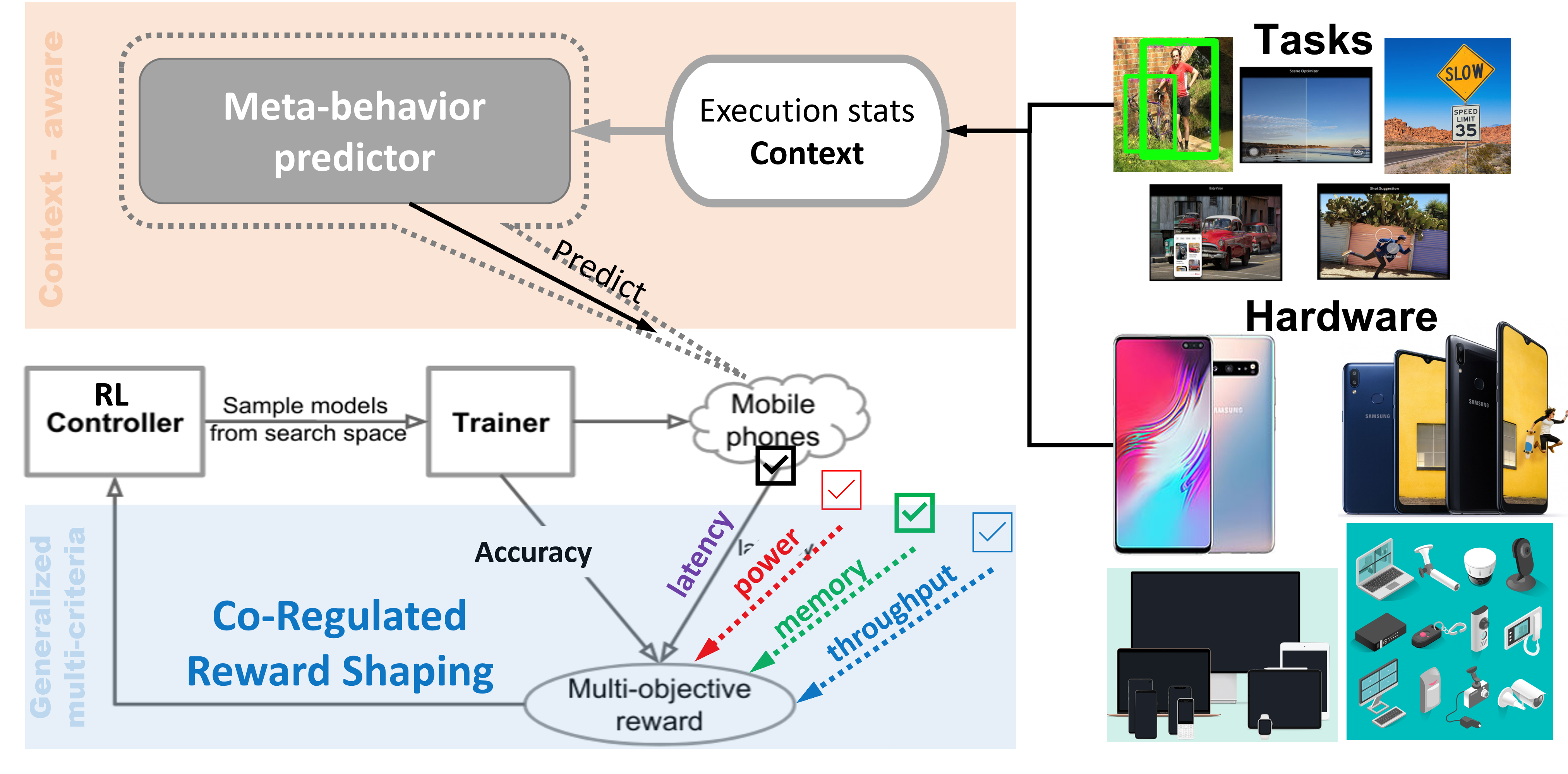}
    \caption{Schematic of \textsc{AutoCoMet} framework}
    \label{fig:architecture}
    \end{minipage}
    \begin{minipage}{0.4\textwidth}
    \centering
    \includegraphics[height=4cm]{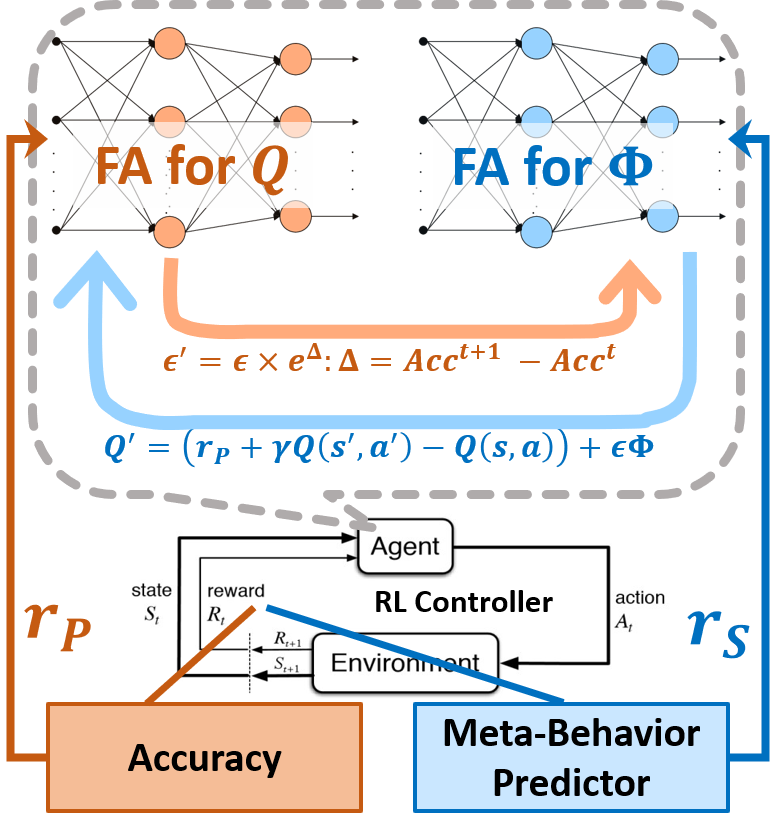}
    \caption{Co-regulated shaping controller}
    \label{fig:shape}
    \end{minipage}
\end{figure*}
Before diving deeper into our proposed framework we define a few terms, 
\begin{definition}[Context]
\label{def:context}
Joint feature space qualifying hardware configuration and task in in factored form is termed as ``Context''. If a hardware configuration $H$ is described by k-tuple $\langle x_H^1, \ldots, x_H^k \rangle$ with features such as processor type, clock frequency, on-chip mem etc. and task/quality requirements $T$ is described by n-tuple $\langle x_T^1, \ldots, x_T^n \rangle$ with features such as text?, multi-entity?, ROI? etc., then Context is a union/concatenation of the 2 tuples $\langle x_C^i \rangle = \langle x_H^1, \ldots, x_H^k \rangle \cup \langle x_T^1, \ldots, x_T^n \rangle$. 
\end{definition}

\begin{definition}[Co-regulated]
\label{def:coreg}
Adapted from the field of social psychology, Co-regulation \cite{fogel2017two} is defined as `continuous unfolding of individual action that is susceptible to being continuously modified by the continuously changing actions of the partner'. In our formalism it refers to how different dimensions of a multi-criteria objective regulate each other's progressive contribution, with the provision for setting \textbf{relative dominance}. [not same as co-learning]
\end{definition}

Figure~\ref{fig:architecture} illustrates the overall architecture of the framework. The RL controller now has a reward  vector with multiple distinct dimensions (factored). For resource constrained devices major components of factored reward are (1) the predictive performance of candidates and (2) the context meta-behavior scores (latency/memory/power/throughput etc.). 
Unlike predictive performance,  measuring context is not straightforward. 
Ex., measuring latency requires deployment/execution of candidate models on actual devices. This is expensive in case of repetitive measurements in NAS controllers. But high fidelity measurements of behavior of candidate architectures is important for learning device ready models. BRP-NAS \cite{dudziak2020brp}, tries to address this by learning complex GCNs to predict measurements but the complexity of this secondary deep model compromise search speed. So, \textsc{AutoCoMet} leverages a high-fidelity shallow predictor model for context. 

Our co-regulated shaping reinforcement can essentially incorporate arbitrary number of meta-behavior components in a smart and generalized way. Meta-Behavior predictor emulates varied pseudo-device/task contexts and output are feeds into the reward vector. Accuracy is the predictive performance component. 
Most existing multi-criteria RL approaches `scalarize' the factored reward vector via weighted combination, which may lead to sub-optimal solutions or slow convergence due to reward sparsity. Also aggressive optimization of hardware metrics may cause deviation from accurate models. \textsc{AutoCoMet} jointly optimizes all the components by allowing them to ``regulate'' one another via a reward-shaping. 

\subsection{Context Meta-Behavior Predictor}
Meta-behavior model is a key factor in context-awareness. \textit{The term "meta" indicates that its purpose is to facilitate learning of the target model via context-aware estimates and it  itself is not our target.} Note context is a hybrid feature space; some features are numerical, some categorical (or ordinal) while one or more are response variables (behavior scores - latency/memory etc.) which are  continuous/real making this a multi-target hybrid regression problem. 

\subsubsection{Feature Design:}
$X=\langle x_A^1,…,x_A^m,x_C^1,\ldots,x_C^{n+k}\rangle$ is the feature space  where, $x_A^i$ are the features that describe candidate DNN architecture, $x_C^j$ signify context features and $\langle y_1,\ldots, y_w \rangle$ are the response variables. Response variables include estimates of latency, memory, power usage, throughput and so on. 
Meta-Behavior function, $y_f=F(X): f\in\{1,\ldots,w\}$ is a function operating on a hybrid feature space, 
which is represented by the piece-wise form more faithfully,
\begin{equation}
\label{eq:yf}
y_f = 
\begin{cases}
 \infty & x_C^i = B\\
 F(X) & \text{otherwise}
\end{cases}
\end{equation}
where $B$ is the set of known values of context for which a given candidate could not be executed at all.
Being a piece-wise function this is not differentiable over the whole space and trivial closed form solution does not exist.
Note that the response variables are assumed independent of each other $y_f \perp y_g| X: f \neq g$ in our multi-target prediction problem. For instance memory heavy models are not always high latency. However, in some scenarios one may imagine some level of dependence such as power consumption may increase in a device that tries to reduce latency. But they are still independent conditioned on the context variables. 

\subsubsection{Predictor Design:}
We choose ensemble regression models as our predictor to deal with the hybrid feature space and piece-wise meta-behavior function. Key idea is that, an ensemble model can faithfully model different parts of the function in different sub-spaces via multiple weak models. 
More specifically, we learn a \textit{bag-of-boosted (B-o-B) regression models} where the outer bag is a Random Regression Forest and each inner weak regression model is built via TreeBoost or Boosted Multi-Layer Perceptron (MLP). Multi-output is handled by constructing one model per target when using boosted trees or having a multi-output MLP otherwise. 

Our empirical evaluations show the validity of our choice and also highlights the difference in performance of the predictor choices. Note that, while MLP is an easier choice because of its ability to directly handle multi-target response, TreeBoost is more interpretable without additional overhead. Note, context values $B$ (Eqn~\ref{eq:yf}) help with informed data sample selection for the `bagging' part of our model. 
Such a predictor generalizes well to wide ranges of hardware/task parameters without evaluating on device during search.

\subsection{Multi-criteria RL Controller with Co-regulated Shaping}
The controller architecture (Figure \ref{fig:shape}), although may seem similar to ProxylessNAS \cite{cai2018proxylessnas}, leverages high-fidelity context meta-behavior predictor as opposed to the naive lookup based latency estimator in ProxylessNAS which is neither generalizable nor nor does it extend to other meta-behavior parameters. Above that we adapt an Q-learning \cite{rummery1994line} based RL update with co-regulated reward-shaping. 

\subsubsection{Decision Process Formulation}We formulate the MDP $\mathfrak{M} = \langle S,A,R,T,\gamma \rangle$ for our problem, where the state space $S$, is naturally factored and same as the predictor feature set $\langle x_A^1,…,x_A^m,x_C^1,…,x_C^{n+k} \rangle$. Action space $A$ is the NAS search space, i.e. the set of possible neural choices. This is customizable and can either be based on some backbone DNN or standardized space such as NAS-Bench \cite{dong2019bench}. Reward vector, $R=\langle r_P,\{r_S\} \rangle$ where $r_P$ is the primary reward signal and $\{r_S\}$ is the set of secondary reward signal. 
This formalism allows our multi-criteria reward to be generalized and extensible. As we will explain later the primary reward component will be the dominant signal in relation to the secondary ones. For proof of concept, we assume $r_P=Accuracy$ and $\{r_S\}=y_f$ (meta-behavior values \textit{Eg: latency}). 
$T$ is the transition function $P(s'|s,a)$ (which is 1 in our case). 
Since we have an infinite horizon problem (DNN can grow infinitely) we assume discount factor $\gamma < 1$.
Reward-shaping formulation regulates the relative contributions of primary and secondary signals seamlessly in an adaptive dynamic fashion, unlike weighted combination approaches.

\subsubsection{Co-regulated reward shaping updates}
{Scalarization} of the reward vector via some weighted additive/multiplicative combination have some major limitations - 
    {(1)} Assumption of common euclidean space for all the components of the multi-criteria reward vector \textit{which is not the case for NAS}
    {(2)} Scalarization essentially computes a mixture distribution, which incorrectly assume all the reward components have same distributional shape. 
    {(3)} The weights attached to the components in the reward vector are {not adaptive to how the learning evolves causing deviation from desired target}. Even simulated annealing cannot mitigate this since the temperature hyperparameter is based on number of iterations. 


Co-regulated reward shaping does not combine/scalarize the components in the reward vector. Instead it allows the secondary reward signal(s) to \textbf{shape} the learning from the primary and leverages the primary to influence the contribution of secondary. We extend \textbf{\textit{PIES (Policy Invariant Explicit Shaping)}} \cite{behboudian2020useful} to achieve the same.
For brevity we assume here that primary reward is {accuracy} $r_P = Accuracy$ and secondary signal is a meta-behavior quantity such as {latency} $r_S = y_f= \{latency\}$, since these are two of the most important quantities that are optimized to obtain device efficient architectures. \ul{It can be easily extended to any arbitrary number of secondary signals (generalized formulation).}

\begin{algorithm}[t!]
\begin{small}
\caption{\textsc{AutoCoMet}}
\label{alg:autometalalgo}
\begin{algorithmic}[1]
\State Pre-process: Context predictor $\mathcal{M} \gets$ \Call{LearnMeta}{$\mathbb{M}$,$\mathbb{E}$}
\Procedure{AutoCoMet}{Dataset $\mathbf{D}$, Predictor $\mathcal{M}$}
\State Initialize: Design space $A \gets \{a_1,\ldots, a_n\}$
\State Initialize: Candidate Neural Architecture $\mathcal{N}$
\While{$\Delta \geq$ Threshold $\tau$}
    \State $a* \gets \underset{a \in A}{\arg\max}~ \pi_{t-1}^*$ \Comment{Choose neural block by policy}
    \State $\mathcal{N}_t \gets \mathcal{N}_{t-1} \Cup a*$ \Comment{$\Cup$ indicates layer addition}
    \State $r_P^t \gets Accuracy =  \Call{Inference}{\mathcal{N}_t(\theta), d\subseteq \mathbf{X}}$
    \State $r_S^t \gets Metabehavior =  \mathcal{M}(\Call{ParseNetwork}{\mathcal{N}_t})$
    \State $\Delta = r_P^{(t)} - r_P^{(t-1)}$
    \State $\delta^\Phi=r_S (s,a)+\gamma \Phi_{(t-1)}(s' , a')-\Phi_{(t-1)}(s,a)$
    \State $\Phi_{t} (s,a)\gets \Phi_{(t-1)} (s,a)+\beta \delta_{(t-1)}^{\Phi}$
    \State Update Function Approximator for $\Phi$: $f(\Theta_S)$
    \State Update $\epsilon$ as per \textbf{Equation~\ref{eq:epsilon}}
    \State $Q'(s,a) \gets Q_t(s,a) + r_P^t + [\gamma^{a} \max_a Q_t(s',a') - Q_t(s,a)]+\epsilon_t \Phi_t (s,a)$
    \State Update Function Approximator for $Q$: $f(\Theta_P)$
    \State Policy $\pi_t \gets softmax(Q_t)$
\EndWhile
\EndProcedure
\algrule
\Procedure{LearnMeta}{DNNset $\mathbb{M}$, Contexts $\mathbb{C}$}
\State Init: Meta-behavior data $\mathbf{Z} = \emptyset$; $Bag(\mathbf{Z}) \gets \emptyset$
\For{Model $\mathfrak{M} \in \mathbb{M}$}
\State $\langle x_A^1,…,x_A^m \rangle_{\mathfrak{M}} \gets$ \Call{ParseNetwork}{$\mathfrak{M}$}
\State $y_f \gets$ Execution stats $\forall C \in \mathbb{C}$
\State $\mathbf{Z}\gets \mathbf{Z}\cup\{\langle x_A^1,…,x_A^m,x_C^1,…,x_C^{n+k}, y_f \rangle\}_{C\in \mathbb{C}}$
\EndFor
\While{NOT Converged}
\State Sample $S \subseteq \mathbf{Z}|B$ \Comment{leveraging $B$ (Eqn~\ref{eq:yf})}
\State BoostRegressor $=$ $XGBoost(S) \text{or} BoostMLP(S)$
\State Add to $Bag(\mathbf{Z}) \gets Bag(\mathbf{Z})~\cup BoostRegressor(S)$
\EndWhile
\State \Return{$Bag(\mathbf{Z})$}
\EndProcedure
\end{algorithmic}
\end{small}
\end{algorithm}

State-space here is intractably large, so we naturally use function approximators to model the value functions. Clearly $r_P$ feeds into the the primary function approximator (FA) $\Theta_P$ and $r_S$ is used to estimate the shaping potential $\Phi$ (via secondary function approximator $\Theta_S$),
	    $\Phi_{(t+1)} (s,a)=\Phi_t (s,a)+\beta \delta_t^{\Phi}$, 
where $\Phi_t$ is the shaping potential, $\delta^\Phi=r_S (s,a)+\gamma \Phi(s' , a')-\Phi(s,a)$ and $\beta$ is learning rate for potential. 
We use shaping potential $\Phi$ to augment the $Q$ updates which uses the primary reward $r_P$,
\begin{align}
    \label{eq:QUpdate}
    \nonumber Q'(s,a)  =  Q(s,a) + r_P  + [\gamma & \max_a Q(s',a') - Q(s,a)]\\
    & +\epsilon \Phi(s,a)
\end{align}
$\gamma = 0.9$ and $\epsilon$ is a decaying trade-off parameter. Till now we have only discussed uni-directional shaping. The formulationof $\epsilon$ is what brings in the co-regulation aspect.  
\begin{equation}
\label{eq:epsilon}
   \epsilon_t=
\begin{cases}
 \epsilon_{(t-1)}\times e^{\Delta} & \epsilon>\text{threshold}\\
 0 & \text{otherwise}
\end{cases}
\end{equation}
where $\Delta$ is the growth in primary reward ($\Delta=Acc_t  -Acc_{(t-1)}$). 
Both FAs are Multi-Layer Perceptrons. 
In original PIES, $\epsilon$ decays at constant rate unlike ours, where its regulated by the primary reward and hence the notion of  \textit{\textbf{`co-regulated'}} (Figure~\ref{fig:shape}), ensuring approximate pareto-optimality. Optimal policy for choice of neural blocks is thus, $\pi_i^*=softmax_{(a\in A)}  (Q_i^* (s,a)+\epsilon \Phi* (s,a))$. \ul{As is clear, the co-regulated trade-off now makes the contribution of the reward components dynamic and adaptive as opposed to the case of scalarization.} 
Finally in case of arbitrary number of meta-behavior quantities like $r_S = \{latency,memory,power,\ldots\} = \{r_S^1,r_S^2,r_S^3,\ldots\}$, we learn multiple potentials $\Phi_1, \Phi_2, \ldots$ and Eqn~\ref{eq:QUpdate} is modified as $Q'(s,a)  =  Q(s,a) + r_P  + [\gamma \max_a Q(s',a') - Q(s,a)]
 +\epsilon^1 \Phi_1(s,a) +\epsilon^2 \Phi_2(s,a) + \ldots$ Also Eqn~\ref{eq:epsilon} is now indexed,
 $$\epsilon_t^i=
\begin{cases}
 \epsilon_{(t-1)}^i\times e^{\Delta^i} & \epsilon>\text{threshold}\\
 0 & \text{otherwise}
\end{cases}$$
where $\Delta^i=r_P^{(t)}-r_P^{(t-1)}$ is computed similarly.
 \ul{Finally also note the we can choose the relative dominance}, and assign some meta-behavior quantity as the primary reward such as $r_P=memory; r_S = \{accuracy, latency, \ldots\}$. Thus \textsc{AutoCoMet} is truly \textbf{generalized} both in terms of choice of relative dominance as well as the set of context meta-behavior variables. It is also \textbf{fast}, as shown in the evaluations, since \ul{co-regulated shaping solves the problem of reward sparsity preventing wastage of search time on irrelevant exploration.}

\subsubsection{\textsc{AutoCoMet} Algorithm}
Algorithm~\ref{alg:autometalalgo} outlines how \textsc{AutoCoMet} leverages predictor and co-regulated shaping. Two different Function Approximators ($\Theta_P$ \& $\Theta_S$) are learned for $Q$ function and shaping potential  $\Phi$ respectively are updated distinctly [Lines 13 \& 16]. Shaping potential is learnable from secondary reward signal (meta-behavior) instead of a constant. All updates to the related variables proceeds as outlined earlier. For brevity, we do not explicitly describe helper functions [Lines 8, 9 \& 23]; eg. \textsc{ParseNetwork}() builds the architectural feature set from a candidate network and \textsc{Inference}() executes a candidate network with test data. \textsc{LearnMeta}() is called only once before search begins and  meta-behavior predictor is learned.

\begin{table*}[t!]
    \centering
    \caption{Comparison of predictive performance, inference latency, architecture size \& learning efficiency of neural architectures learned by \textsc{AutoCoMet} against baselines (Top-1 Accuracy in Appendix). We report IOU for segmentation}
    \scalebox{0.80}{
    \begin{tabular}{p{0.10\linewidth}|l|c|c|c|c|c|c|c|c|c|c|c|c}
    \hline
      \multirow{2}{0.0pt}{\textbf{Task}} & \multirow{2}{0.0pt}{\textbf{Model}}  & \multicolumn{3}{c|}{\textbf{Top-5 Acc / IOU \%}} & \multicolumn{3}{c|}{\textbf{Inf Latency (milliseconds)}} & \multicolumn{3}{c|}{\textbf{Size\%}} & \multicolumn{3}{c}{\textbf{Search Time (GPU hrs)}}\\
      \cline{3-14}
      & & 1 & 2 & 3 & 1 & 2 & 3 & 1 & 2 & 3 & 1 & 2 & 3\\
      \hline
      \hline
        \multirow{8}{0.10\linewidth}{{Classification (ImageNet)}} & MobileNetV2 \textit{[Manual]} & 91.2 & 91.2 & 91.2 & 113.2 & 256.2 & 300.5 & 100 & 100 & 100 & -- & -- & --\\
        & ProxylessNAS (MNAS-Net) & 88.5 & 88.5 & 85.8 & 76.8 & 193.5 & 235.6 & 70 & 70 & 70 & 5.8 & 6 & 7.6\\
        & NAS+Predictor & 90.24  & 90.24 & 89.9 & \textbf{50.28} & \textbf{129.25} & \textbf{130.8} & 70 & 70 & 60 & 10.8 & 10.8 & 13\\
        & \textit{\textsc{AutoCoMet}} & 90.48 & 90.48 & 90.2 & \textbf{50.28} & \textbf{129.56} & \textbf{130.8} & 70 & 70 & 60 & \textbf{3} & \textbf{3} & \textbf{3.6}\\
        \cline{2-14}
        & InceptionV3 \textit{[Manual]} & 94.49 & 94.3 & 94.39 & 346.2 & 445.8 & 505.5 & 100 & 100 & 100 & -- & -- & --\\
        & ProxylessNAS (MNAS-Net) & 90.5 & 90.5 & 89.0 & 234.8 & 278.4 & 280.5 & 80 & 80 & 65 & 11.4 & 12 & 15.8\\
        & NAS+Predictor & 92.6 & 92.6 & 92.6 & 111.81 & 154.54 & 176.8 & 80 & 75 & 62 & 25 & 22.5 & 26.3\\
        & \textsc{AutoCoMet} & 92.6 & 93.2 & 92.9 & 111.67 & 156.2 & 178.5 & 80 & 75 & 62 & \textbf{11} & \textbf{8.8} & \textbf{13.6}\\
        \hline
        \multirow{4}{0.10\linewidth}{Segmentation (COCO)} & MobileNetV2+Decoder & 89.9 & 89.9 & 89.9 & 45.2 & 57.6 & 76.5 & 100 & 100 & 100 & -- & -- & --\\
        & ProxylessNAS (MNAS-Net) & 83.5 & 83.5 & 83.8 & 30.8 & 32.9 & 35.1 & 64.5 & 60.5 & 60 & 9.6 & 11.2 & 11.2\\
        & NAS+Predictor & 85.8  & 85.78 & 85.4 & 26.8 & 39.5 & 35.8 & 70 & 65.5 & 68.2 & 15.6 & 13.8 & 14\\
        & \textsc{AutoCoMet} & 88.4 & 86.8 & 87.1 & $\approx$\textbf{13} & \textbf{15.3} & \textbf{30.8} & 70 & 62.5 & 62.5 & \textbf{3.5} & \textbf{3.5} & \textbf{4}\\
        \hline
    \end{tabular}}
    \label{tab:globalComp}
\end{table*}

\section{Experimental Evaluation}
Empirical validation of automated architecture learning is not straightforward. Hence, our evaluation aims to answer the following –
    \textbf{[Q1.]} Is \textsc{AutoCoMet} able to generate architectures having better predictive performance or higher efficiency? 
    \textbf{[Q2.]} Is the meta-behavior predictor effective? 
    and 
    \textbf{[Q3.]} How fast is our novel co-regulated controller? 
\begin{table}[h!]
    \centering
    \caption{Comparison of meta-behavior predictors}
    \scalebox{0.82}{
    \begin{tabular}{c|c|c|c}
    \hline
       \textbf{Model}  & \textbf{R2} & \textbf{RMSE} & \textbf{Time}\\
       \hline
       \hline
        B-o-B & \textbf{0.9417} & \textbf{1.32} & 31ms\\
        XGBoost & {0.91} & {1.354}& 29.9ms\\
        MLP & 0.8748 & 1.36 & 28.8ms\\
        SVR & 0.8253 & 1.5142 & \textbf{199.5$\mu$s}\\
        \hline
        B-o-B & \multicolumn{2}{c|}{Acc. = $97.2\pm0.6$}& 31ms\\
        GCN (BRPNas) & \multicolumn{2}{c|}{Acc. = $96.9\pm0.78$}& 93ms\\
        \hline
    \end{tabular}}
    \label{tab:metaPerf}
\end{table}

\subsection{Experimental setting}
We are primarily interested in the computer vision domain and thus we evaluate on such tasks. Baselines are, 1) ProxylessNAS (related MNASNet) itself and 2) Vanilla expert designed state-of-the-art networks such as MobileNetV2 \cite{sandler2018mobilenetv2}, InceptionV3. 
We implement execution stats collection pipeline and co-regulated RL controller in python 3.5 with pyTorch v1.4. All architecture search experiments are run on 2 core Nvidia 1080Ti GPU system (both our approach and baselines).

\noindent\textbf{Domains: }We evaluate on 2 distinct tasks (data sets), namely \textit{ImageNet} \cite{imagenet_cvpr09} for object classification and \textit{COCO dataset} \cite{lin2014microsoft} for `segmentation'. In COCO dataset we change contrast and brightness and then find the object mask to create the segmentation labels. 
The displayed results are averaged over 10 training runs. 
Meta-predictor is trained with the execution stats dataset that we collect from various mobile devices with varying hardware configurations super-sampled with SMOTE technique \cite{chawla2002smote}. Layerwise inference latency, memory/power usage data was collected for multiple DNN models including VGG, Squeezenet, RCNN, MobilenetV1/V2, GoogleNet, AlexNet etc. using 
publicly available Samsung Neural SDK \cite{samsungneural}. We capture the following attributes - `Type, Kernel Size, Stride, Padding, Expansion Ratio, Idskip, Channels, Height, Width, Input Volume, Output Volume, Execution time, Cores, Compute Units, Memory, Clock Freq. and Memory B/w'


\subsection{Experimental Results}
We conducted 3 types of experiments. To show that \textsc{AutoCoMet} can design DNN architectures that are both accurate as well as optimally efficient on an ecosystem of mobile devices we compare the vanilla manually designed state-of-the-art architectures to the ones that were generated by \textsc{AutoCoMet}. Also, we compare architectures generated by related methods, ProxylessNAS (which itself is based on MNASNet search space and optimization) as well as NAS+Predictor (which is \textsc{AutoCoMet} w/ meta-behavior predictor but w/o co-constrained shaping). We execute the learned models on 3 types of mobile hardwares - `Type 1' (flagship devices w/ high memory b/w + CPU/GPU/Neural-Processors.), `Type 2' (Mid-tier device w/ reasonable memory b/w + CPU/GPU) and `Type 3' (low-tier devices w/ low memory and CPU). 
To analyse the fidelity of our meta-behavior predictor we perform ablation studies with different predictors and show that our B-o-B predictor is most effective.
Finally, to show that our novel co-constrained RL controller results in faster convergence of the architecture search, we present as learning curves of the cumulative returns in \textsc{AutoCoMet} against ProxylessNAS (which subsumes other related approaches such as MNASNet). 

\subsubsection{Quality of generated architecture \textbf{[Q1]}}

Table~\ref{tab:globalComp} shows the performance/efficiency on classification and segmentation tasks. As described earlier we repeat the same experiments on different device types (Types 1, 2 \& 3). We show results against 2 popular classification models (MobilenetV2 and Inception V3) and an expert designed segmentation model MobilenetV2+Decoder with Deconv layers stacked after MobilenetV2 base (similar to UNet). 
We report Top-5 Accuracy (classification) and Intersection-Over-Union/IOU (segmentation), average inference latency, ratio of size of the learned architectures and search-time (Top1 reported in Appendix\footnote{For Appendix refer to supplementary file})

Across all device types \textsc{AutoCoMet} learns architectures with predictive performance equal or close to state-of-the-art, while the inference latency as well as the size of the learned architecture is almost always significantly lower. Results for Type 3 are even more significant -- \textsc{AutoCoMet} learns smaller and more efficient architectures. On segmentation task we achieve $~13ms$ latency on flagship device which is the most ambitious till date. Note that ProxylessNAS (related MNASNet) although does produce reasonably efficient architectures (closer to ours), suffers in accuracy unlike \textsc{AutoCoMet}. This supports our conjecture that our approach tries to achieve the ideal level of relative dominance in the joint optimization. 

Additionally we see that `NAS+predictor', though produces effective and efficient architectures, yet has higher search time. After analysing the intermediate outputs we did find that simple combination-based approach of multi-criteria RL suffers due to reward sparsity. \textsc{AutoCoMet} on the other hand handles it really well due to the shaping-based strategy. 

\subsubsection{Fidelity of meta-behavior predictor \textbf{[Q2]}}
\label{sec:metaresult}

Accurate prediction of context-sensitive execution meta-behavior is important for the shaping potential. To understand the relative effectiveness we compare our \textit{B-o-B} against both ensemble and traditional regression model, such as XGBoost, Support Vector Regression (SVR) and Multi-Layer Perceptron (MLP). Table~\ref{tab:metaPerf} shows our ensemble model \textit{B-o-B} outperforming the others (as expected given the hybrid feature space). It is even marginally than GCN predictor in BRP-NAS, while being much faster. Note that GCN predictor is very limited since it has boolean response (good/bad). 
Response plots (Appendix) also show marginal `unexplained variance' resulting in high $R^2$ scores.

\begin{figure}[t!]
    \centering
    \includegraphics[width=0.60\columnwidth]{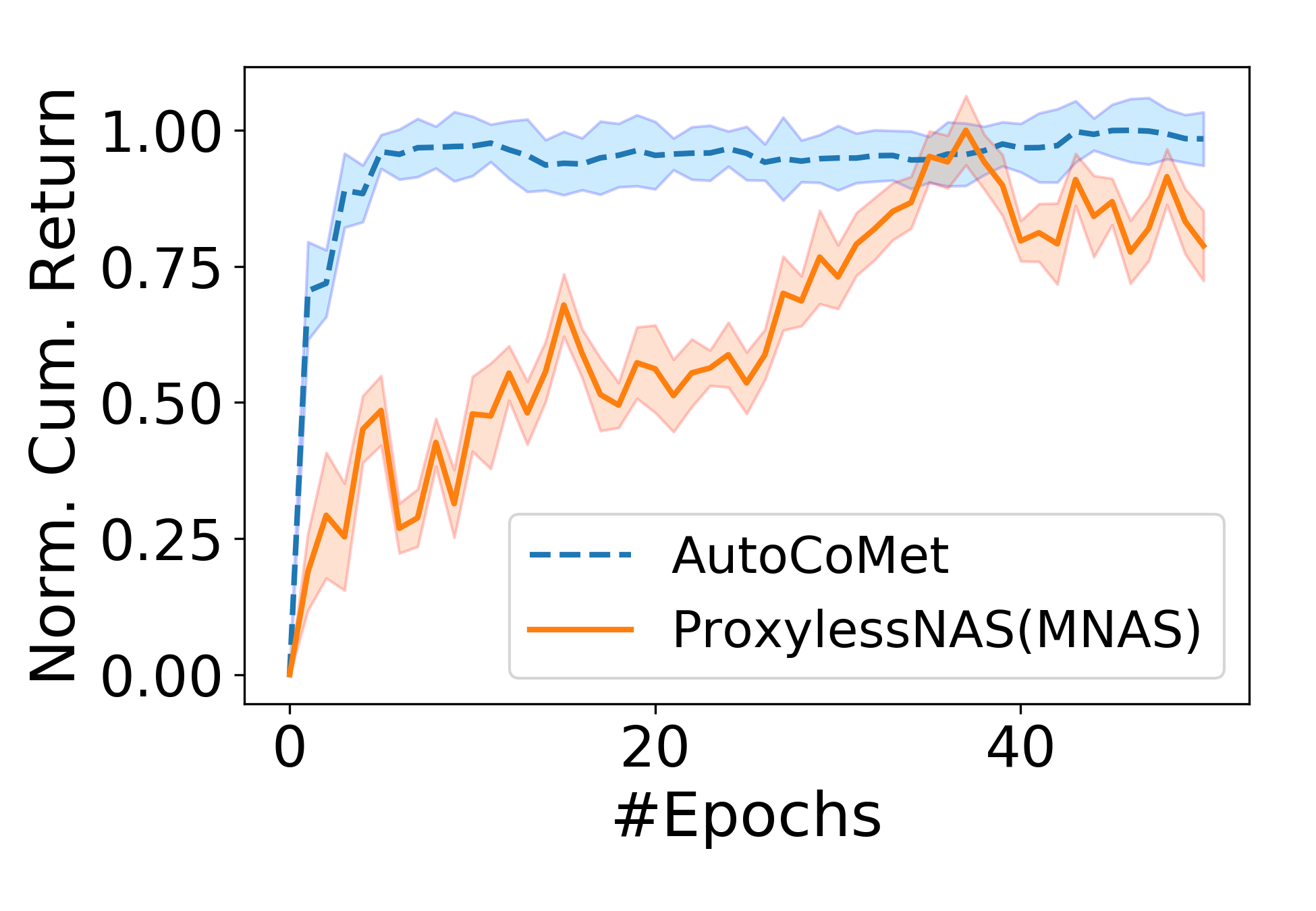}
    \caption{Learning curve on Imagenet classification task} 
    \label{fig:LC1}
\end{figure}

\subsubsection{Rapid Search / Sample efficiency \textbf{[Q3]}}
Figure~\ref{fig:LC1} shows the learning curves with respect to $[0,1]$ normalized cumulative returns (averaged) of the RL controllers of \textsc{AutoCoMet} against ProxylessNAS (MNAS), for Type-1 hardware configuration. We observe that \textsc{AutoCoMet} plateaus out by 4-5 epochs while ProxylessNAS gradually climbs to similar cumulative returns by $\approx 40$ epochs. This ablation study aims to show the difference in the core RL strategies (beyond the the final performance shown in Table~\ref{tab:globalComp}). These learning curves provide insight into the convergence dynamics of the core RL controllers in the different NAS methods. This supports our claim the our co-regulated mechanism implicitly takes care of of reward sparsity, hence converging early. 
Note that each epoch takes consumes 0.2 GPU hours on an average. 
The learning curve with respect to the Segmentation task is provided in the Appendix.

\section{Conclusion}
We propose our NAS/AutoML framework \textsc{AutoCoMet} that leverages high fidelity meta-predictor of device behavior for a wide range of devices/tasks (contexts) and uses a novel co-regulated shaping RL controller to generate optimal DNN architectures for an ecosystem of contexts. We also show how our controller leads to faster convergence which is a very important feature for application developers interested in using the proposed framework to learn suitable DNN architectures for their AI-driven mobile applications even without access to cloud or high-powered computational resources. The context-awareness and co-regulated shaping formulation of the multi-criteria feedback 
provides a generalized and extremely fast architecture learning framework irrespective of context and quality requirements.  It also implicitly adaptive and allows for customizing the relative conservativeness in the multi-objective optimization.  

One caveat however, is that careful training of the mete-behavior predictor is important for robustness. 
Capturing more exhaustive device behavior data is am important exercise that we aim to focus on next. Though, as outlined earlier, our formalism allows for learning over arbitrary combination of quality metrics in a principled manner, yet experimentation with other device behavior metrics (memory, power etc.) and adaptation to IoT (Internet of Things) ecosystem of connected devices are open challenges. 

\bibliographystyle{IEEEtran}
\bibliography{biblio}



\section*{Supplementary material}

# Supplementary Material

## 1 APPENDIX

### 1.1 Meta-Behavior Predictor Ablation Study

Above are the response plots for all the meta-predictors.

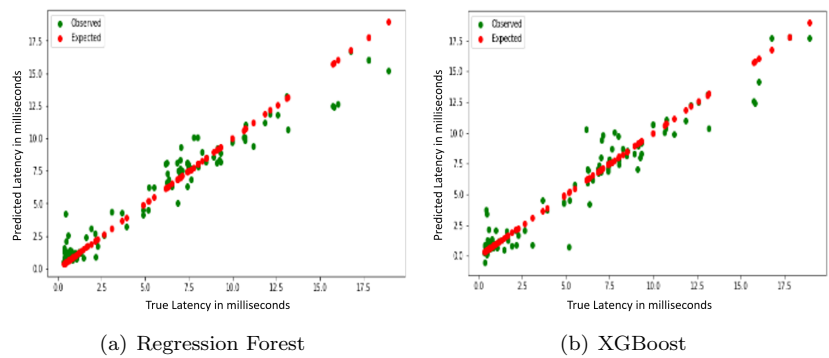

(a) Regression Forest  (b) XGBoost

Figure 1: Response plots for Ensemble models - Regression Forest & XGBoost. x-axis = true latency values and y-axis = predicted latency estimates.

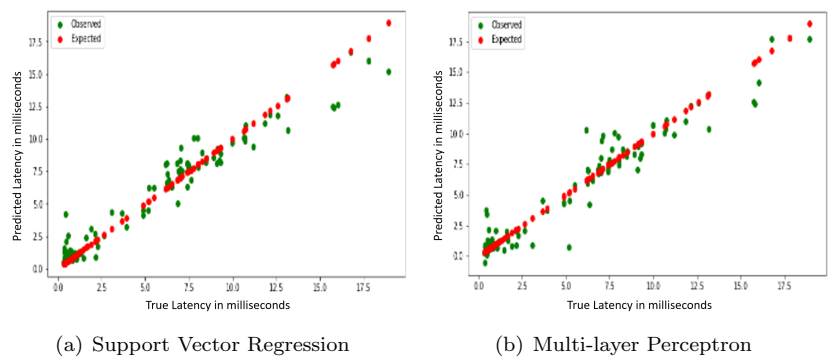

(a) Support Vector Regression  (b) Multi-layer Perceptron

Figure 2: Response plots for Non-Ensemble models - SVR & MLP. x-axis = true latency values and y-axis = predicted latency estimates.

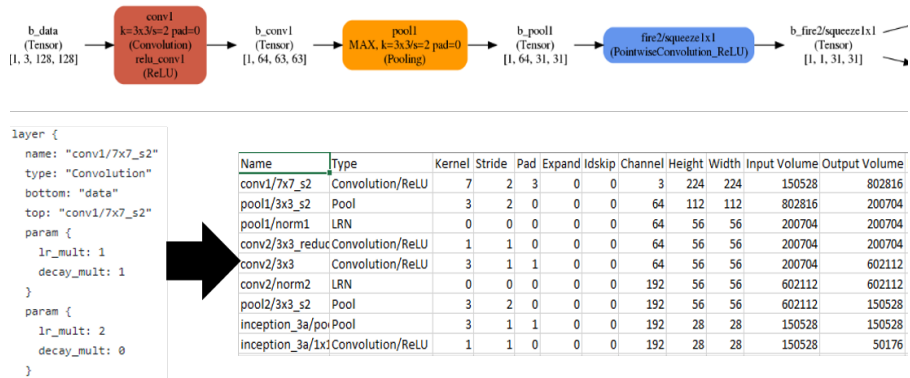

Figure 3: Parsing models and stats into feature sets

### 1.1.1 Execution Stats Collection:

Execution stats data collection pipeline for training of meta-predictor, as described in the main manuscript, is done via profiling the execution inference performed by different types of Deep Neural Networks on different (android) mobile devices with varying hardware configurations. The data dumps taken via the Android Logcat are then parsed and the layerwise execution stats are then obtained such as execution time (latency), memory usage. An in-house battery monitoring tool is used to measure model-wise power usage. To get layer-wise power usage we repeatedly execute one layer at a time and then measure battery usage.

After all the execution stats are collected it is parsed to create the feature set comprising hardware, task and architecture parameters as discussed in the main paper.

## 1.2 Other results

Below table 1 shows the Top-1 accuracies of the classification models. The main paper only contains Top-5 accuracies due to space constraints and the fact that both types accuracies are positively correlated and leads to teh same insight.

Below Figure 4 shows the normalized cumulative return plot as a learning curve for the segmentation task between naive RL based NAS controller like ProxylessNAS (and/or MNASNet). The main paper only contains the plot with respect to the imagenet classification task.

Table 1: Comparison of predictive performance, inference latency, architecture size & learning efficiency of neural architectures learned by AutoCoMet against baselines – Top-1 Accuracy

| Task | Model | Top-1 Accuracy % | | | Inf Latency (milliseconds) | | |
|---|---|---|---|---|---|---|---|
| | | 1 | 2 | 3 | 1 | 2 | 3 |
| Classify (ImageNet) | MobileNetV2 | 72.19 | 72.19 | 72.19 | 113.2 | 256.2 | 300.5 |
| | ProxylessNAS (MNASNet) | 69.14 | 68.92 | 66.2 | 76.8 | 193.5 | 235.6 |
| | NAS+Predictor | 71.2 | 71.15 | 70.11 | **50.28** | **129.25** | **130.8** |
| | AutoCoMet | 72.12 | 72.05 | 71.95 | **50.28** | **129.56** | **130.8** |
| | InceptionV3 | 77.69 | 77.69 | 77.69 | 346.2 | 445.8 | 505.5 |
| | ProxylessNAS (MNASNet) | 74.65 | 74.65 | 73.04 | 234.8 | 278.4 | 280.5 |
| | NAS+Predictor | 77.5 | 77.14 | 75.45 | 111.81 | 154.54 | 176.8 |
| | AutoCoMet | 77.6 | 77.62 | 77.5 | 111.67 | 156.2 | 178.5 |

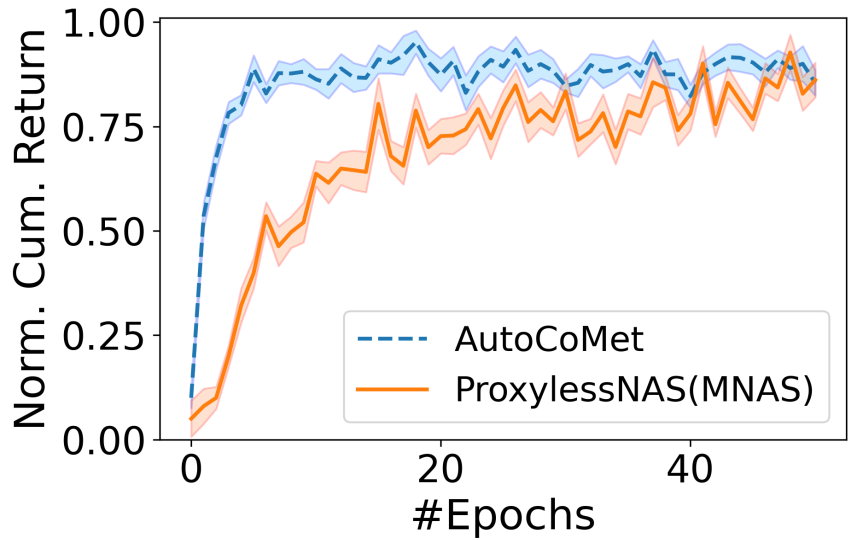

Figure 4: Learning curve for the **Segmentation** task


\end{document}